\documentclass{article}

\usepackage[preprint]{nips_2018}

\usepackage[utf8]{inputenc} 
\usepackage[T1]{fontenc}    
\usepackage{hyperref}       
\usepackage{url}            
\usepackage{booktabs}       
\usepackage{amsfonts}       
\usepackage{nicefrac}       
\usepackage{microtype}      
\usepackage{amsmath,amssymb,bbm}
\usepackage{graphicx}
\usepackage{subcaption}
\usepackage{xspace,xcolor,natbib}
\usepackage{multicol}
\usepackage{enumitem}

\usepackage{algorithm,algpseudocode}

\newcommand{\states}{\ensuremath{\mathcal{S}}}
\newcommand{\actions}{\ensuremath{\mathcal{A}}}
\newcommand{\transitions}{\ensuremath{\mathcal{T}}}
\newcommand{\rewards}{\ensuremath{\mathcal{R}}}
\newcommand{\observations}{\ensuremath{\mathcal{O}}}
\newcommand{\tree}{\ensuremath{\mathcal{X}}}

\newcommand{\simulator}{\ensuremath{\textsc{Query-Model}}}
\newcommand{\oracle}{\textsc{Oracle}\xspace}
\newcommand{\fullenv}{\textsc{Full-Env}\xspace}
\newcommand{\rewardsonly}{\textsc{Rewards-Only}\xspace}
\newcommand{\drqn}{\textsc{DRQN}\xspace}

\definecolor{ForestGreen}{RGB}{11,102,35}

\definecolor{LakeBlue}{RGB}{104, 120, 201}

\title{Variational Inference for Data-Efficient Model Learning in POMDPs}

\author{
  Sebastian Tschiatschek\\
  Microsoft Research \\
  \texttt{sebastian.tschiatschek@microsoft.com} \\
  \And
  Kai Arulkumaran\thanks{Part of this work was done while Kai was a Research Intern at Microsoft Research Cambridge.} \\
  Imperial College London \\
  \texttt{ka709@ic.ac.uk} \\
  \And
  Jan St\"{u}hmer \\
  Microsoft Research \\
  \texttt{t-jastuh@microsoft.com} \\
  \And
  Katja Hofmann \\
  Microsoft Research \\
  \texttt{katja.hofmann@microsoft.com} \\
}

\begin{document}

\maketitle

\begin{abstract}
Partially observable Markov decision processes (POMDPs) are a powerful abstraction for tasks that require decision making under uncertainty, and capture a wide range of real world tasks. Today, effective planning approaches exist that generate effective strategies given black-box models of a POMDP task. Yet, an open question is how to acquire accurate models for complex domains. In this paper we propose DELIP, an approach to model learning for POMDPs that utilizes structured, amortized variational inference. We empirically show that our model leads to effective control strategies when coupled with state-of-the-art planners. Intuitively, model-based approaches should be particularly beneficial in environments with changing reward structures, or where rewards are initially unknown. Our experiments confirm that DELIP is particularly effective in this setting.
\end{abstract}


\section{Introduction}

Reinforcement learning (RL) is a form of machine learning where one or more agents learn by trial and error from interactions with an environment. RL is a very general learning framework with applications ranging from video and other game play \citep{tesauro1995temporal,Mnih2015} to robotics \citep{kober2013reinforcement,quillen2018deep}, (visual) dialog \citep{singh2000reinforcement,das2017learning}, health \citep{edwards2013temporal,kidzinski2018learningtorun}, and a host of other domains.

A key challenge in RL is \emph{data-efficient} learning. State of the art approaches to, e.g., learning to play Atari games, are trained on 10s to 100s of millions of samples to achieve competitive performance \citep{machado2017revisiting,vanseijen2017hybrid}. Relying on vast amounts of data limits applicability of these types of approaches to domains where data can be obtained relatively easily, e.g., in simulations or video games. However, even when accurate simulations are available, the computational cost of training these approaches is immense.

In this paper, we address the problem of data-efficient learning in \emph{partially observable Markov decision processes} (POMDPs). The POMDP setting is a problem formulation that is particularly relevant for many real-world applications, where agents observe a local signal (e.g., a first-person view of a 3D world, or a set of diagnostics in a health care application). To deal with partial observations, agents need to explicitly or implicitly consider interaction history to reason about possible underlying states of the world that may have generated current observations \citep{singh1994learning}. This exacerbates the data-efficiency problem: a learning agent now has to collect and learn from enough data to learn how its behavior should depend on, possibly long, interaction sequences, instead of just individual state observations (as is the case in the fully observable MDP setting that is more commonly addressed in RL and planning).

We propose to leverage recent advances in structured variational inference \citep{krishnan2015deep,krishnan2017structured,fraccaro2016sequential} to learn generative models of POMDP dynamics and rewards. Recent progress in this area has lead to effective learning algorithms that effectively capture complex dynamics by parameterizing posterior distributions using deep neural networks. The result is a versatile, data-driven approach to learning dynamics models in a wide range of environments --- without requiring domain expertise often required in previous approaches. We empirically show that the learned models result in effective control strategies when coupled with state-of-the-art black-box planning algorithms \citep{silver2010POMCP}. We further compare to a a recent off-policy, model-free algorithm designed for POMDPs \citep{hausknecht2015deep} that directly learns a strategy from observations using recurrent networks, and demonstrates competitive performance. Model-based approaches are hypothesized to be particularly advantageous in environments with consistent dynamics but changing or initially unknown rewards structures --- our experiments confirm that our approach is particularly data-efficient in such a setting.

The remainder of the paper is structured as follows. 
In Section~\ref{sec:background} we introduce our notation and review relevant background on POMDPs and amortized variational inference.
In Section~\ref{sec:model} we describe our proposed generative model of the POMDP environment, DELIP (\emph{Data Efficient model Learning in POMDPs}), and detail how we use it for planning.
In our experiments in Section~\ref{sec:experiments} we investigate how well our model can be learned from data on an actual RL task and investigate its sample efficiency relative to contemporary baseline models.
Related work is reviewed in Section~\ref{sec:related}.
Finally, we conclude our paper with a discussion and an outlook on future work in Section~\ref{sec:discussions}.



\section{Notation and Background}
\label{sec:background}

\subsection{Partially Observable Markov Decision Processes (POMDPs)}

\begin{figure}[ht]
  \centering
  \includegraphics{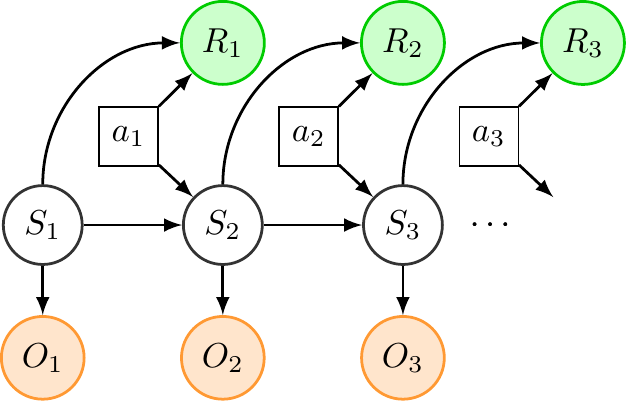}
  \caption{POMDP. At each timestep $t$, the state $s_t$ depends stochastically on the previous state $s_{t-1}$ and an external action $a_{t-1}$ chosen by an agent. The agent receives a stochastic observation $o_t$ as well as a deterministic reward $r_t$. Finally, the agent will decide on its next action $a_t$. In traditional RL algorithms for POMDPs, a belief is held over the current state, based on the previous actions, belief, rewards and observations \citep{kaelbling1998planning}.}
  \label{fig:pomdp}
\end{figure}

POMDPs are defined as 7-tuples $(\states, \actions, \transitions, \rewards, \Omega, \observations, \gamma)$, where $\states$ is a set of states, $\actions$ is a set of actions, $\transitions$ is a set of state-conditional transition probabilities, $\rewards$ is a set of state-conditional reward distributions, $\Omega$ is set of possible observations, $\observations$ is a set of state-conditional observation probabilities and $\gamma \in (0, 1]$ is the discount factor.
POMDPs can be considered as \emph{controlled hidden Markov models}, as visualized in Figure~\ref{fig:pomdp}.
They consist of time-dependent latent variables $s_t \in \states$, (partial) observations $o_t \in \observations$ of the state $s_t$, rewards $r_t \in \mathbb{R}$ and actions $a_t \in \actions$. The actions are provided by an agent interacting with the environment according to the POMDP. State transitions, observations and rewards are generated as $s_t \sim \transitions(s_t|s_{t-1}, a_{t-1})$, $o_t \sim \observations(o_t|s_t)$ and $r_t = \rewards(r_t|s_t, a_t)$, respectively, where, to simplify notation, we overload the notation of the state-dependent distributions.

The goal of an agent interacting with the environment is to learn and execute a policy $\pi$ that maximizes the expected cumulative discounted reward $\mathbb{E}[R(\pi)] = \mathbb{E}[\sum_{t=1}^{\infty} \gamma^{t-1} r_t]$, where the expectation is over the randomness of the environment and randomness of the policy $\pi$. The expected future performance of a policy can also be quantified from a given state --- as expressed by the value function $V^\pi(s_t) = \mathbb{E}[\sum_{k=1}^{\infty} \gamma^{k-1} r_{t+k-1}|s_t]$.

Approaches for learning the policy $\pi$ can be broadly categorized into \emph{model-free} approaches and \emph{model-based} approaches. Both types of approaches have long research traditions, but much of the work has been pursued in parallel, with little connection between these strands. Model-free approaches directly learn to maximize rewards without modeling the underlying environment, making them versatile and effective with little prior knowledge \citep{singh1994learning,hausknecht2015deep}. Model-based approaches derive effective policies when accurate dynamics models are available, yet assuming accurately specified models of complex real-world tasks is often unrealistic \citep{silver2010POMCP,katt2017learning}.

We are interested in data-efficiency and therefore focus on model-based approaches. We address the key question of how models can be learned effectively with minimal assumptions about the problem domain (to avoid model mis-specification). Our approach draws on potential solutions from several current strands of research. In particular, the combination of variational inference, which provides a theoretical foundation for inferring latent variable models, and deep learning, which uses powerful function approximators, allows us to learn models that can be applied to 
POMDP problems with very high levels of partial observability.


\subsection{Variational Inference}

Variational inference can be used to learn latent variable models by maximizing a lower bound of the marginal probability density $p(o) = \int p(o|s)p(s) ds$, where $s$ are latent variables and $o$ are observed variables. The term  $p(o|s)$ is the likelihood of the data given the latent variables $s$ and, in the case of neural networks, can take the form of a parameterized model. 

Instead of dealing with the prior $p(s)$ directly, variational autoencoders (VAEs) infer $p(s)$ using the posterior $p(s|o)$ \citep{kingma2013auto,rezende2014stochastic}. As the true form of the posterior distribution $p(s|o)$ is unknown, variational inference turns the problem of inferring latent variables into an optimisation problem by approximating the true posterior distribution with a variational distribution $q(s|o)$, which takes the form of a simpler distribution such as a fully factorized Gaussian, and then minimising the Kullback-Leibler (KL) divergence between $q(s|o)$ and $p(s)$. As the KL divergence is nonnegative and minimised when $p$ is the same as $q$, the training objective for VAEs is known as the variational or evidence lower bound (ELBO):
\begin{align}
\mathcal{L}(q; x) 
                  &= \mathbb{E}_{s \sim q(s)}[\log p(o|s)] - D_{\text{KL}}[q(s|o)\Vert p(s)]. \label{eqn:elbo} 
\end{align}

VAEs also utilize amortized inference \citep{gershman2014amortized}, reparameterized variables, and stochastic gradient variational Bayes \citep{kingma2013auto,rezende2014stochastic}. VAEs consist of a generative model $p(o|s; \theta)$ with parameters $\theta$ and an inference model $q(s|o; \psi)$ with variational parameters $\psi$ that can be trained using stochastic gradient descent.


\subsubsection{Structured Variational Inference}


For a POMDP, the joint probability of states $s$, observations $o$, and rewards $r$ conditioned on the actions $a_t$ from the initial timestep $t=1$ to the end of a trajectory of length $T$ is:
\begin{align}
\begin{split}
p(o_{1:T}, r_{1:T}, s_{1:T} \mid a_{1:T}) &= p(o_1 | s_1)p(r_1 | s_1, a_1)p(s_1) \\
&\quad\prod_{t=2}^T p(o_t|s_t)p(r_t|s_t, a_t)p(s_t|s_{t-1}, a_{t-1}). \label{eq:model}
\end{split}
\end{align}
For this factorisation, the true posterior for a single latent variable $s_t$ depends on all future observations, future rewards and actions, i.e.\ $p(s_t \mid s_{1:T}, o_{1:T}, r_{1:T}, a_{1:T}) = p(s_t \mid o_{t:T}, r_{t:T}, a_{t:T})$ \citep{krishnan2015deep,krishnan2017structured,fraccaro2016sequential}.
Consequently, one can assume a corresponding factorization for the variational posterior and use recurrent neural networks (RNNs) for summarizing future observations, rewards and actions, cf.\ Section~\ref{sec:model-learning}.

\section{Model \& Planning}
\label{sec:model}

\subsection{Model}
\label{sec:model-learning}

We consider models for POMDPs in which all distributions, i.e.\ the prior state distribution, the state transitions, the observation probabilities and the reward probabilities, are parameterized by neural networks. That is, our model takes the form of Equation~\eqref{eq:model}, where:
\begin{align*}
p(s_1) &= \mathcal{N}(\mu_0, \sigma_0), \\
p(s_t \mid s_{t-1}) &= \mathcal{N}(s_t \mid \mu^s(s_{t-1}), \sigma^s(s_{t-1})), \\
p(o_t \mid s_t) &= \mathcal{N}(o_t \mid \mu^o(s_t), \sigma^o(s_t)), \\
p(r_t \mid s_t) &= \mathcal{N}(o_t \mid \mu^r(s_t), \sigma^r(s_t)),
\end{align*}
with $\mu_0, \sigma_0$ as parameters and where $\mu^s(\cdot), \sigma^s(\cdot)$, $\mu^o(\cdot)$, $\sigma^o(\cdot)$, $\mu^r(\cdot)$ and $\sigma^r(\cdot)$ are parameterized neural networks. We learn these parameters from data in the form of $N$ trajectories \mbox{$\mathcal{D} = \{ (o^{(1)}_{1:T}, r^{(1)}_{1:T}, a^{(1)}_{1:T}), \ldots, (o^{(N)}_{1:T}, r^{(N)}_{1:T}, a^{(N)}_{1:T}) \}$} using structured, amortized variational inference.

We use a variational posterior of the form \mbox{$q(s_{1:T} | o_{1:T}, r_{1:T}, a_{1:T})=\prod_{t=1}^T q(s_t \mid s_{t-1}, o_{t:T}, r_{t:T}, a_{t:T}),$} where each $q(s_t \mid \cdots)$ is parameterized by an RNN summarizing the varying length future observations, rewards and actions from time $t$ to $T$ and a neural network mapping the summary of the RNN together with the state $s_{t-1}$ to the parameters of a 4-state Gaussian mixture model over states $s_t$.

For training we jointly optimize the ELBO in Equation~\eqref{eqn:elbo} over the variational posterior $q(s | \cdots)$ and the generative model given by $p(s_1), p(s_t \mid s_{t-1}), p(o_t \mid s_t), p(r_t \mid s_t)$. We make use of the reparametrization trick to reduce the variance of gradient estimates \citep{kingma2013auto,rezende2014stochastic}.

\subsection{Planning}
\label{sec:planning}

Here we describe the algorithm we use for planning, using our learned model of the POMDP. We base our planner on partially observable Monte-Carlo planning (POMCP), a form of Monte-Carlo tree search (MCTS) with upper confidence bounds \citep{kocsis2006bandit} that uses the full history instead of individual states in order to apply MCTS to POMDPs \citep{silver2010POMCP}. The use of Monte-Carlo methods is important in the context of data-efficient learning in POMDPs, as their sample complexity is independent of the state or observation spaces, and only on the underlying difficulty of the POMDP \citep{kearns2000approximate}. We describe POMCP in Algorithm~\ref{alg1}.
In contrast to the original POMCP algorithm, we do not rely on a potentially prohibitive simulator, but instead use our learned model for rollouts.

The POMCP planner is invoked by $\textsc{Search}(P(S))$, where $P(S)$ is the current belief about the true state of the POMDP (initially, we assume a uniform distribution over states).
The planner builds up a search tree $\tree$, in which each node corresponds to a hypothetical state of the environment.
For each node, the algorithm keeps track of $N(s',a)$, i.e.\ the number of times action $a$ was taken in state $s'$, an estimate $V(s',a)$ of the value of taking action $a$ in state $s'$ and a list of successor states.
During execution, the planner traverses the search tree if and expands it at its leaf nodes.
Rollouts with a random policy $\pi_{\textnormal{rollout}}$ are used to initially estimate the value of states.
While traversing the search tree, the algorithm balances exploration and exploitation to decide on whether to execute an action that looks promising or an action that has been executed only infrequently \citep{kocsis2006bandit}.

In POMCP, the term $\simulator(s, a)$ stands for invoking the simulator to replicate the execution of action $a$ in state $s$.
In contrast, we sample the next state $s'$, observation $o$ and reward $r$ for taking action $a$ in state $s$ from our trained model.
After having decided on taking a particular action, i.e.\ after $\textsc{Search}(P(S))$ has terminated, we receive an actual observation and reward information from the environment and use these to update the distribution $P(s)$ over states $s$ using Bayesian filtering (again using the model).

\begin{algorithm}
	\caption{POMCP with a simulator}
	\label{alg1}
    \begin{algorithmic}
    \Require Parameter $c$ for balancing exploration and exploitation, number of simulations $T_\textnormal{sim}$ for chosing next action, discount factor $\gamma$
    \vspace{-3mm}
    \begin{multicols}{2}
    \Procedure{Search}{$P(S)$}
      \For{$t=1,\ldots,T_{\textnormal{sim}}$}
        \State $s \sim P(s)$
        \State \textsc{Simulate}(s, 0)
      \EndFor
      \State \Return {$\arg\max_a V(s,a)$}
    \EndProcedure%
    \\   
    \Procedure{Rollout}{$(s, \textrm{depth})$}
      \If{$\gamma^\textrm{depth} \leq \epsilon$}
        \State \Return{$0$}
      \EndIf      
      \State $a \sim \pi_\textrm{rollout}(s)$
      \State $(s', o, r) \sim \simulator(s,a)$
      \State \Return $r + \gamma \textsc{Rollout}(s',depth+1)$
    \EndProcedure
    \\ \\ \\
    \Procedure{Simulate}{$(s, \textrm{depth})$}
      \If{$\gamma^\textrm{depth} \leq \epsilon$}
        \State \Return{$0$}
      \EndIf
      \If{$s \in \tree$}
        \For{$a \in \actions$}
          \State $T(s,a) \leftarrow (N_\textrm{init}(s,a), V_\textrm{init}(s,a), \emptyset)$
          \State \textsc{Rollout}(s, \textrm{depth})
        \EndFor
      \EndIf
      \State $a \leftarrow \arg\max_{b} V(s,b) + c \sqrt{\frac{\log N(s)}{N(s,a)}}$
      \State $(s', o, r) \sim \simulator(s,a)$
      \State $R \leftarrow r + \gamma \textsc{Simulate}(s',depth+1)$
      \State $N(s) \leftarrow N(s) + 1$
      \State $N(s,a) \leftarrow N(s,a) + 1$
      \State $V(s,a) \leftarrow V(s,a) + \frac{R-V(s,a)}{N(s,a)}$
      \State \Return $R$
    \EndProcedure
    \end{multicols}
    \vspace{-3mm}
    \end{algorithmic}
\end{algorithm}


\section{Experiments}
\label{sec:experiments}

\subsection{Experimental Setup}

\begin{figure}
  \centering
  \begin{subfigure}[t]{0.6\linewidth} 
    \includegraphics[width=1.\textwidth]{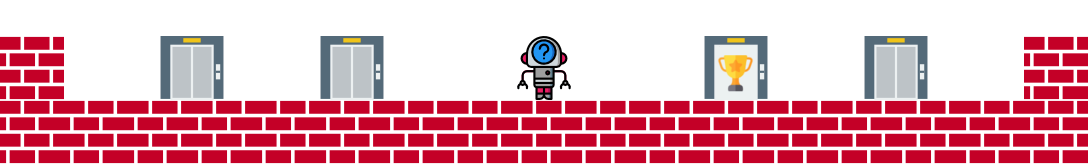}
    \caption{Task illustration}
    \label{fig:robot-task-ill}
  \end{subfigure}\\
  \begin{subfigure}[t]{0.5\linewidth}
    \centering
	 \includegraphics[width=0.9\textwidth]{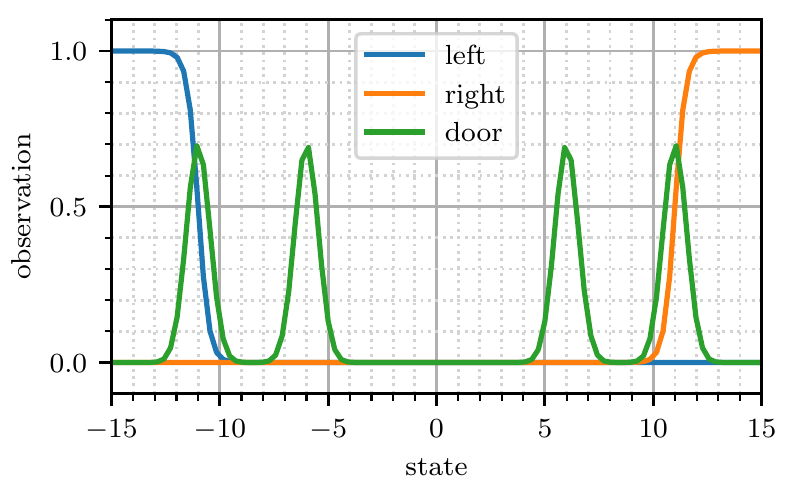}
    \caption{Signals}
    \label{fig:robot-task-signals}
  \end{subfigure}%
  \begin{subfigure}[t]{0.5\linewidth} 
    \centering
    \includegraphics[width=0.9\textwidth]{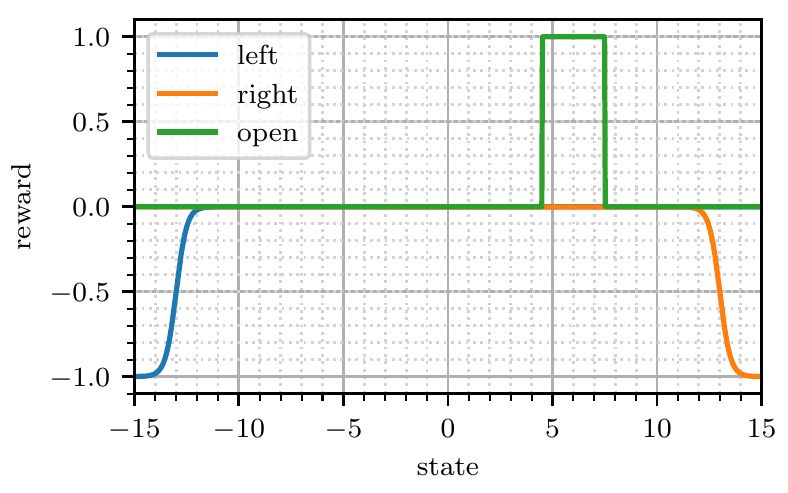}
    \caption{Rewards}
    \label{fig:robot-task-rewards}
  \end{subfigure}
  \caption{Robot navigation task. \emph{(\subref{fig:robot-task-ill}}) Illustration of the task---a robot with limited observation capabilities can gather positive rewards by opening the second but rightmost door. (\subref{fig:robot-task-signals}) The robot does not observe its actual location, but only signals indicating whether it is at a boundary of the environment or in front of a door. (\subref{fig:robot-task-rewards}) The robot receives a positive reward if it opens the correct door.}
  \label{fig:robot-task}
\end{figure}

In our experiments we consider an autonomous navigation task in which a robot has to find and open a door given limited observation capabilities.
This navigation task is inspired by~\cite{porta2005robot} and illustrated in Figure~\ref{fig:robot-task-ill} and described below.

{\bfseries Navigation task.} 
In each episode, the robot is randomly placed in the space according to the Gaussian distribution with zero mean and a standard deviation of $5$.
If the robot is placed outside the \emph{boundaries} of the world ($-15$ and $15$), its position is clipped to the boundary.
In any position, the robot can take any action in $\actions = \{\textnormal{left}, \textnormal{right}, \textnormal{open}\}$.
The goal of the robot is to find and open the second but last door from the right.
Opening the correct door yields a reward of $+1$, trying to move outside of the boundaries of the world results in negative rewards.
Taking the action ``left'' moves the agent to the left by unit and taking the action ``right'' moves the agent to the right by unit.
The robot cannot observe its actual position but is only provided signals indicating whether it is at the left or right limits of the space and whether it is in the front of a door (but there is no signal indicating in front of which door it is).
For successfully solving this task, the robot has to identify its position in space, navigate to the correct door and open it.
The observations and rewards of the robot are shown in Figures~\ref{fig:robot-task-signals} and~\ref{fig:robot-task-rewards}, respectively.
It is important to note that any approach using only information from the current observation (or a short sequence of observations) will not be effective for this task, as the information that can be gathered from multiple observations has to be combined to identify the robot's position from the observations.

\subsection{Learning Environment Models}

\begin{figure}
  \centering
  \begin{subfigure}[t]{0.48\textwidth}
    \centering
    \includegraphics{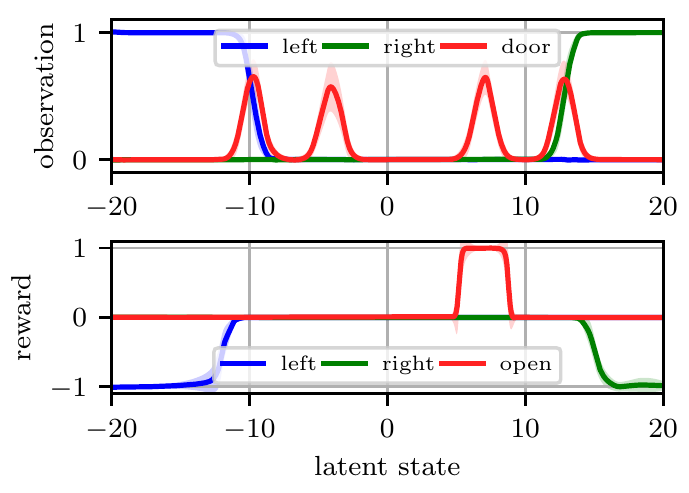}
    \caption{Successful learning}
    \label{fig:success}
  \end{subfigure}\hspace{2mm}%
  \begin{subfigure}[t]{0.48\textwidth}
    \centering
    \includegraphics{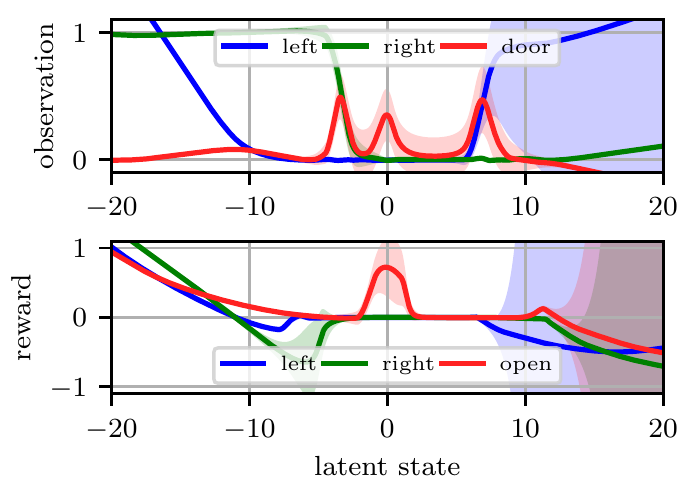}
    \caption{Missing door}
    \label{fig:missing-door}
  \end{subfigure}%
  \caption{Success and failure cases for learning the full environment model. \emph{(\subref{fig:success})} Successfully learned environment model. The observations and rewards predicted from the model closely resemble the ground-truth. \emph{(\subref{fig:missing-door})} Unsuccessfully learned model. The observation model cannot predict all four doors correctly. This manifests in high variance of the predictions.}
  \label{fig:learned-env-model}
\end{figure}

Before turning to evaluating the generative models learned via variational inference as described in Section~\ref{sec:model-learning} for reward maximization, we showcase success and failure cases that frequently occurred when learning the model, cf.\ Figure~\ref{fig:learned-env-model}.
For the transition model we used a linear model; the observation probabilities and the reward probabilities are parameterized by three layer neural networks with ReLU activation functions and $100$ neurons each.
The prior distribution is a normal distribution with learned mean and variance. 
For the approximate posterior model, we use a bidirectional LSTM with $10$ hidden units.
For learning we used Adam~\citep{kingma2014adam} with a batch size of $100$, and trained the models for up to $10000$ epochs, i.e.\ $10000$ passes through the dataset, or until convergence. 
To stabilize learning, we initially clamped the variance of all distributions\footnote{We clamped the log-variance to $-3$ for the first $1{,}000$ epochs. After these $1{,}000$ epochs, the variance was not constrained.}.
In addition, we used the following non-uniform sampling strategy to correct for the class imbalance caused by having sparse positive rewards.
For each trajectory in the dataset we computed the cumulative reward.
We uniformly quantized the cumulative rewards of all trajectories into 5 bins.
Finally, when sampling a mini-batch, we selected an equal number of trajectories from each bin, thereby ensuring that we observe positive and negative reward accumulation behaviour in each mini-batch.

These results are all for the same model architecture, and only hyper-parameters like learning rate used for optimization, weight decay or whether gradient clipping was used, were varied.\footnote{We also observed large variance of the obtained results for different random seeds.}
A successfully learned environment model is visualized in Figure~\ref{fig:success}.
The predictions of the model are very close to the ground-truth, cf.\ Figure~\ref{fig:robot-task}.
In Figure~\ref{fig:missing-door}, we show a common issue arising when learning the environment, i.e.\ a missing door.
An agent using this model for planning typically achieves poor performance as it infers its position incorrectly.

\subsection{Planning}
We now turn to evaluating the learned environment models for planning.
We are interested in (a) validating that we can use recent advances in variational inference for learning a generative model of the environment that is useful for planning, and (b) comparing our approach with other contemporary classical approaches for learning and planning in POMDPs in terms of data efficiency. 

{\bfseries Learning POMDPs.}
We consider learning of POMDPs in two different variants:
\begin{enumerate}[leftmargin=*]
  \item \emph{Learning the full environment (\fullenv).} Here we want to learn a full environment model, i.e.\ the model for transitions, observations and rewards.
  \item \emph{Learning rewards only (\rewardsonly).} Here we are interested in learning only the reward function, given a pretrained model of the environment\footnote{We did not use the sampling strategy described in the previous section for pretraining.} without a reward model.
    Such settings arise naturally in scenarios in which agents have to solve different tasks in the same basic environment.
\end{enumerate}
For both variants we collect data from the environment by executing a random policy which chooses between all possible actions uniformly, for 100 steps (i.e.\ episodes of length 100).
The collected trajectories consist of observations and rewards for the first variant, and rewards only for the second variant.
We learn the POMDPs from different numbers of collected episodes and and evaluate the learned models when used for planning.
For the second variant, we pretrain an model for the environment without observing the reward, using a dataset consisting of $8{,}000$ episodes.


{\bfseries Models and Baselines.}
We compare the following approaches:
\begin{itemize}[leftmargin=*]
  \item DELIP. This implements POMCP using our trained model as a simulator for the environment, cf.\ Section~\ref{sec:planning}.
      To keep track of the state distribution with particles, we discretize the state-space into intervals of size $0.1$ 
      For each planning step, we perform $2{,}000$ simulations, rolling out simulations for $30$ steps.
      
  \item \drqn~\citep{hausknecht2015deep}. This can be considered as a variant of deep Q-networks \cite{Mnih2015} in which the last layer of hidden units in the deep Q-network is connected with an RNN (we use an LSTM~\citep{hochreiter1997long} in our experiments).
  This enables the deep recurrent Q-network (\drqn) to introduce temporal dependencies on previous observations and Q-values.
	As an off-policy algorithm, the \drqn is naturally more sample-efficient than competitive on-policy algorithms that utilise RNNs \cite{mnih2016asynchronous}. In addition, the DRQN's use of experience replay \citep{lin1992self} can be seen as placing it between model-free and model-based methods \citep{vanseijen2015deeper}.
    For our DRQNs, we use two fully connected layers with $50$ neurons each and ReLU activation functions, to process the input.
    The output of the second layer is connected over time via LSTM cells with 50 neurons each.
    Finally, the output of the hidden units of the LSTM cells is further processed by a a fully connected layer with 50 neurons and ReLU activations and fed through a final linear layer predicting Q-values for each action.
    We trained the DRQNs for $10{,}000$ epochs using Adam, using a mini-batch size of 100 trajectories, a learning rate of $0.001$ and updating the target Q-network every 100 epochs.
    For the \rewardsonly scenario, we pretrain the DRQN by training it on 3 auxillary tasks, namely maximizing the observation of the signal ``left'', maximizing the observation of the signal ``right'' and maximizing the observation of the signal ``door''.
    
  \item \oracle. This baseline implements POMCP~\citep{silver2010POMCP}. It has access to a perfect simulator of the environment but does not have access to the true state of the environment the agent is actually interacting with, i.e.\ \oracle can perform simulations in the environment which follow the true transition and observation probabilities.
As with DELIP, we use particles to model the state distribution and discretize the state-space intro intervals of size $0.1$.
        Given infinite simulation time per planning step (with a sufficiently fine-grained quantization of the state-space), this baseline provides an upper bound on the achievable performance.
        For each planning step, we perform $2{,}000$ simulations, rolling out simulations for $30$ steps.
\end{itemize}

\begin{figure}
  \centering
    \centering
    \includegraphics{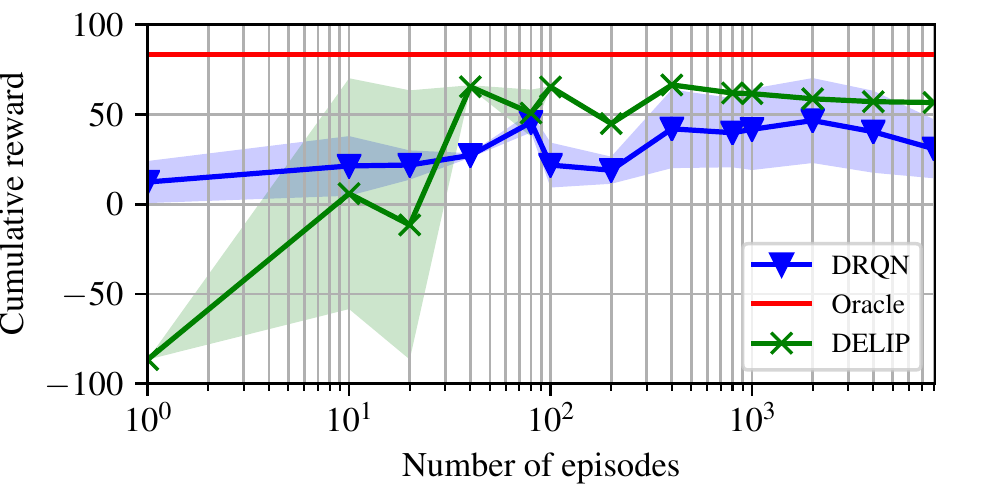} 
    \label{fig:rewards-only}
\caption{Performance over number of training samples for learning the reward distribution on top of a pre-trained environment model. DELIP outperforms the pre-trained model-free baseline for all but very small sample sizes. For these small sample sizes, DELIP does not reliably learn that opening a door can lead to a positive rewards. In contrast, DRQN, due to its pre-training with auxiliary tasks, typically learns policies that invoke the ``open'' action upon observing that the robot is in front of a door.}
  \label{fig:planning}
\end{figure}

{\bfseries Results.}
We compare the average performance in terms of cumulative regret achieved by DELIP, DRQNs and \oracle on 100 episodes of interaction with the environment, each consisting of 100 time steps and using $\gamma=1$.
Our main results on data-efficiency of DELIP is that if the full environment model is learned, DELIP and DRQNs perform on par (plot omitted due lack of space). For the case in which only the reward distribution on top of a pre-trained environment model is learned, DELIP performs favorably, as shown in Figures~\ref{fig:planning}.
We observe that DELIP outperforms the \drqn for most dataset sizes.
Thus, there is a regime in which using the model-based approach is advantageous compared to the model-free approach, even when the latter is pre-trained on relevant auxiliary tasks. 

\section{Related Work}
\label{sec:related}

Learning and planning in POMDPs are key research challenges with a long tradition of theoretical and empirical work. Here we summarize a selection of works that are most closely related to the present paper. In addition, we build on research in deep state space models, which we also review, especially where these have been applied in an RL context.


Model-free learning in POMDPs is a main focus of RL, where a behavior policy is learned without explicitly modeling environment dynamics. Pioneering work by \cite{singh1994learning} established that learning optimal deterministic memoryless policies could perform arbitrarily worse than either optimal stochastic policies or policies with memory. Recently, \cite{hausknecht2015deep} demonstrated competitive performance of model-free reinforcement learning approaches in high-dimensional, partially observable settings, with recurrent models capable of maintaining memory. Here we compare to their approach, DRQN. Promising extensions of this work consider explicit forms of memory together with attention mechanisms that learn to access these \citep{oh2015action} - understanding data efficiency compared to these is an important direction for future work.


Research on POMDPs originated in operations research, and initially focused on deriving optimal policies from given models, e.g., using variants of dynamic programming \citep{sondik1978optimal,kaelbling1998planning}. An insightful survey of early work is by \cite{aberdeen2003revised}. 
A strong focus is on finding scalable solutions. Scalability is key because even POMDPs with few latent states can induce a very high-dimensional belief states that render exact methods intractable. A breakthrough in scalable planning for POMDPs was achieved by \cite{silver2010POMCP}, who demonstrated scalability to problems with $10^56$ belief states. Here, we use their approach, POMCP, as a black-box planner.


Work on planning in POMDPs was able to progress while abstracting from the problem of model acquisition. Assuming an accurate domain model is given before the start of interaction between agent and environment is unrealistic in most practical applications. One line of work investigates how to relax assumptions on model accuracy, e.g., by assuming a parametric model which allows for online updates as new experience is collected. This setting is captured in the BA-POMDP framework \citep{ross2008bayes,katt2017learning}. Compared to the present work, significantly more domain knowledge would be required to instantiate the parametric models used in this line of work. In contrast we demonstrate flexible model learning with minimal assumptions.

Our work builds on recent progress in variational inference for learning deep state space models \citep{krishnan2015deep,krishnan2017structured,fraccaro2016sequential}. Our model is most closely related to the Deep Kalman Filter (DKF) proposed by \citeauthor{krishnan2015deep} The DKF has strong representational power and can model rich partially observable environments while at the same time enabling an easy integration with POMCP, as all information about the environment  is represented in the latent state variables. In other state-space models, e.g.\ those additionally including RNNs, the integration with POMCP poses additional challenges as the RNN state and the state of the latent variables have to be considered jointly.

To the best of our knowledge, our work is the first to demonstrate that models learned through modern variational inference techniques coupled with state-of-the-art black-box planning approaches can result in competitive behavior policies in POMDP settings. Previous work demonstrated variational inference to learn models for POMDP settings \citep{moerland2017learning,fraccaro2018generative} but without closing the loop to learn or derive behavior policies. The use of (deterministic) transition models was proposed for augmenting data \citep{kalweit2017uncertainty} and observations \citep{racaniere2017imagination,buesing2018learning} for learning model-free policies.

\section{Conclusion}
\label{sec:discussions}

We addressed the problem of data-efficient learning in POMDP settings. Our proposed approach draws from several recent advances and can flexibly learn dynamics and reward models that, as we demonstrate, lead to effective control strategies. This result opens up exciting research directions towards more flexible and data-efficient learning in POMDPs with minimal prior assumptions.
Future work will seek to understand advantages and trade-offs between modern approaches to model learning, e.g., comparing to recently proposed 
autoregressive models \citep{van2017neural}.
Moving beyond black-box planning approaches, recent progress in ``learning to plan'' \citep{henaff2017prediction} could enable principled and scalable model-based approaches that are learned end-to-end.




%


\bibliographystyle{apalike}
\bibliography{references}

\end{document}